\documentclass[11pt]{article}

\usepackage[final]{acl}

\usepackage{times}
\usepackage{latexsym}
\usepackage[T1]{fontenc}
\usepackage[utf8]{inputenc}
\usepackage{microtype}
\usepackage{inconsolata}
\usepackage{graphicx}
\usepackage{subcaption}
\usepackage{amsmath,amssymb,amsthm}
\usepackage{booktabs}
\usepackage{multirow}
\usepackage{algorithm}
\usepackage{algpseudocode}
\usepackage{xcolor}
\usepackage[capitalise]{cleveref}

\newcommand{\KL}{D_{\mathrm{KL}}}
\newcommand{\E}{\mathbb{E}}
\newcommand{\Vocab}{\mathcal{V}}

\newcommand{\student}{\pi_\theta}
\newcommand{\teacher}{\pi_T}
\newcommand{\SLOT}{\textsc{KAT}}

\newcommand{\given}{\mid}

\title{Escaping the KL Agreement Trap in On-Policy Distillation}

\author{
  Haoran Xin\textsuperscript{1} \quad
  Anhao Zhao\textsuperscript{3,4} \quad
  Ying Sun\textsuperscript{1$\ast$} \quad
  Jin Li\textsuperscript{1} \quad
  Xiaoyu Shen\textsuperscript{4} \quad
  Hui Xiong\textsuperscript{1,2$\ast$} \\
  \textsuperscript{1}Thrust of Artificial Intelligence, The Hong Kong University of Science and Technology (Guangzhou) \\
  \textsuperscript{2}Department of Computer Science and Engineering, \\
  The Hong Kong University of Science and Technology, Hong Kong SAR, China \\
  \textsuperscript{3}The Hong Kong Polytechnic University \quad
  \textsuperscript{4}Eastern Institute of Technology, Ningbo \\
  \texttt{hxin883@connect.hkust-gz.edu.cn} \quad
  \texttt{sunyinggilly@gmail.com} \quad
  \texttt{xionghui@ust.hk}
}

\begin{document}
\maketitle

\renewcommand{\thefootnote}{\fnsymbol{footnote}}
\footnotetext[1]{Corresponding authors.}
\renewcommand{\thefootnote}{\arabic{footnote}}
\setcounter{footnote}{0}

\begin{abstract}
On-policy distillation (OPD) provides dense token-level supervision by asking a teacher to score student-generated rollouts.
However, when the student drifts into an unrecoverable prefix, the teacher may locally agree with the degraded state, producing low reverse KL but little corrective training signal.
We identify this persistent regime as a \emph{low-KL agreement trap}.
Further analyses show that tokens during and after such traps produce less useful supervision signals.
We propose \SLOT{} (\emph{\underline{K}L \underline{A}greement \underline{T}rap Termination}), an online OPD termination rule that detects persistent low-KL agreement with a dynamic training-adaptive threshold.
By filtering weak supervision from degenerate agreement, \SLOT{} improves avg@$k$ accuracy by 2.66\% and pass@$k$ by 3.43\% across four mathematical benchmarks, while reducing average rollout length by 59.73\%.
\end{abstract}

\section{Introduction}
\label{sec:introduction}

Post-training has become a decisive stage in the modern large language model (LLM) pipeline, where capabilities such as multi-step reasoning~\cite{chen2025reasoning,guo2025deepseek,kumar2025llm} and tool use~\cite{qi2025webrl,wang2025ragen} are largely shaped.
A central challenge in this stage is how to transfer the capabilities of strong but expensive models to smaller, deployable ones.
LLM distillation~\cite{xu2024survey} addresses this by using teacher models to provide richer supervision than ground-truth labels alone, improving smaller models' reasoning ability.

Within this landscape, {on-policy distillation} (OPD)~\cite{li2026rethinking,lu2025onpolicydistillation,agarwal2024policy} has recently emerged as a promising recipe for training students on states closer to their own inference-time behavior, and now appears in the post-training pipelines of several flagship models~\citep{deepseekai2026deepseekv4,xiao2026mimo,yang2025qwen3}.
At its core, OPD lets the student sample its own rollouts and asks the teacher to score every visited state with a dense, token-level signal.
This mitigates the {exposure bias} that off-policy distillation~\cite{kim2016sequence} typically suffers from, while yielding far higher sample efficiency than sparse outcome-reward reinforcement learning~\citep{shao2024deepseekmath,schulman2017proximal}.

\begin{figure*}[t]
    \centering
    \includegraphics[width=\linewidth]{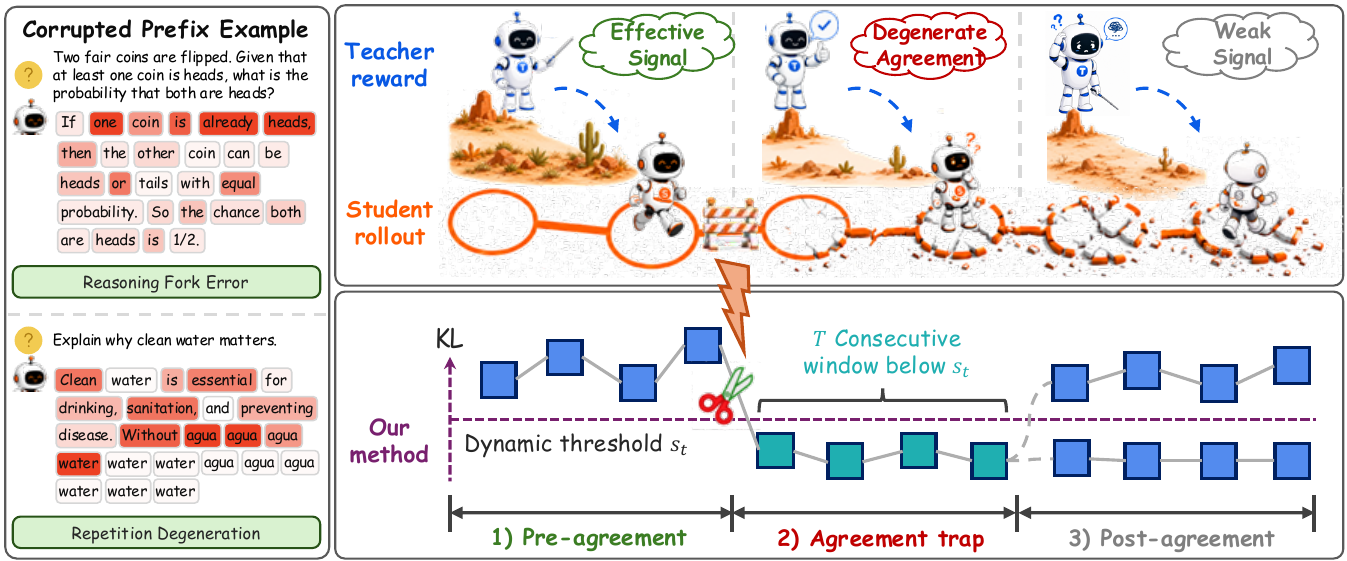}
    \caption{Overview of \SLOT{}. A student rollout can drift into a corrupted prefix, e.g., an incorrect reasoning fork or repetitive degeneration, where the teacher locally agrees with the degraded state. Darker blocks denote higher KL. Sustained low KL indicates a {low-KL agreement trap}, after which the teacher provides weak corrective supervision. \SLOT{} reuses OPD's reverse KL, tracks a sliding-window statistic with a training-adaptive threshold, and terminates rollouts once persistent low-KL agreement is detected, focusing updates on informative pre-trap tokens.}
    \label{fig:framework}
    \vspace{-4mm}
\end{figure*}

However, existing OPD methods often treat the full student rollout as useful supervision.
We identify an underexplored failure mode: as rollouts progress, the student may drift into a corrupted prefix, such as committing an early reasoning error or entering a repetitive degeneration pattern, as illustrated in \Cref{fig:framework}.
Conditioned on this degraded state, the teacher's token-level distribution may carry little corrective information, instead producing continuations that are plausible under the erroneous prefix and therefore close to the student’s distribution~\cite{lu2025onpolicydistillation}. In this case, low KL can reflect local teacher–student agreement in a corrupted state rather than progress toward a correct solution.
%
This gives low teacher–student KL two contrastive identities: \textbf{benign agreement}, where the student follows a correct reasoning path, and \textbf{degenerate agreement}, where the teacher locally follows a corrupted prefix with little corrective signal, providing weak training signal and wasting computation on uninformative tokens.
This makes the central challenge distinguishing when low KL marks degenerate supervision that can misdirect and dilute OPD updates.

Our key observation is that degenerate agreement leaves a temporal signature beyond token-level low KL: after a certain rollout depth, the teacher–student KL can remain low across consecutive local regions. We call this persistent regime a \textbf{low-KL agreement trap}. 
This persistence is crucial for distinguishing degenerate agreement from benign agreement, since both can exhibit low KL locally, but only the former indicates sustained agreement on a corrupted prefix.
Our analyses show that supervision in the agreement region is weak, and the subsequent post-agreement suffix does not recover effective teacher signal.
Thus, once a rollout enters the agreement trap, continuing generation is unlikely to provide useful supervision and may dilute earlier learning signals.

Building on this diagnosis, we propose \textbf{\SLOT{}} (\emph{\underline{K}L \underline{A}greement \underline{T}rap Termination}), a simple training-time rollout termination method for OPD (\cref{fig:framework}).
\SLOT{} reuses the teacher--student reverse KL already computed by OPD and monitors whether it remains low across consecutive local regions.
Once this persistent low-KL pattern is detected, \SLOT{} terminates the rollout online.
The stopping threshold is dynamic, calibrated from recent rollout statistics and adapted to the evolving KL scale throughout training.
By removing suffix tokens after degenerate agreement, \SLOT{} shifts the effective OPD training signal toward more informative prefixes and improves rollout efficiency without introducing a new loss, auxiliary model, or more than $O(1)$ bookkeeping per token.

Empirically, across four mathematical benchmarks and two student scales, \SLOT{} improves avg@$k$ accuracy by 2.66\% and pass@$k$ by 3.43\%, while reducing average rollout length by 59.73\%.
Particularly, it also outperforms random termination and fixed-prefix truncation, showing that low-KL agreement traps provide an effective online signal for filtering weak supervision rather than merely shortening rollouts.
Our contributions are:
\begin{itemize}
    \item We identify {low-KL agreement traps} in OPD, where persistent teacher--student agreement indicates weak rather than useful supervision.

    \item We analyze that supervision remains low-quality after entering the trap, including in the post-agreement suffix.

    \item We propose \SLOT{}, an adaptive online termination rule that removes uninformative suffixes without changing the OPD objective.
\end{itemize}

\section{Preliminaries}
\label{sec:preliminaries}

\subsection{Notation}
\label{sec:notation}

Let $x$ denote a prompt and $\hat y = (\hat y_1, \dots, \hat y_T)$ a student rollout sampled from the student policy $\student(\cdot \given x)$.
We write $h_t = (x, \hat y_{<t})$ for the prefix at step $t$, and use $p_t(v) = \student(v \given h_t)$ and $q_t(v) = \teacher(v \given h_t)$ for the student and teacher next-token distributions, respectively.
The token-level reverse KL divergence is
\begin{equation}
\label{eq:kl}
    d_t \;=\; \KL\!\left(p_t \,\|\, q_t\right) \;=\; \sum_{v \in \Vocab} p_t(v) \log \frac{p_t(v)}{q_t(v)}.
\end{equation}

\subsection{On-Policy Distillation Objective}
\label{sec:opd-objective}

We adopt the standard OPD objective \citep{agarwal2024policy,li2026rethinking}:
\begin{equation*}
\label{eq:opd-loss}
\begin{aligned}
\mathcal{L}_{\mathrm{OPD}}(\theta) \;=\; \E_{x, \hat y \sim \student(\cdot \given x)} \Bigg[ \sum_{t=1}^{T} & \KL\!\big( \student(\cdot \given x, \hat y_{<t}) \\[-4pt]
& \quad \big\| \teacher(\cdot \given x, \hat y_{<t}) \big) \Bigg].
\end{aligned}
\end{equation*}
The expectation is taken over student-generated trajectories: supervision is dense (every generated token receives a teacher signal) and is anchored on the student's own state distribution.
This combination is what gives OPD its sample-efficiency advantage over outcome-only RL methods, and its robustness advantage over off-policy KD that trains only on a static dataset.
\begin{figure*}[t]
    \centering
    \begin{subfigure}[c]{0.49\linewidth}
        \centering
        \includegraphics[width=\linewidth]{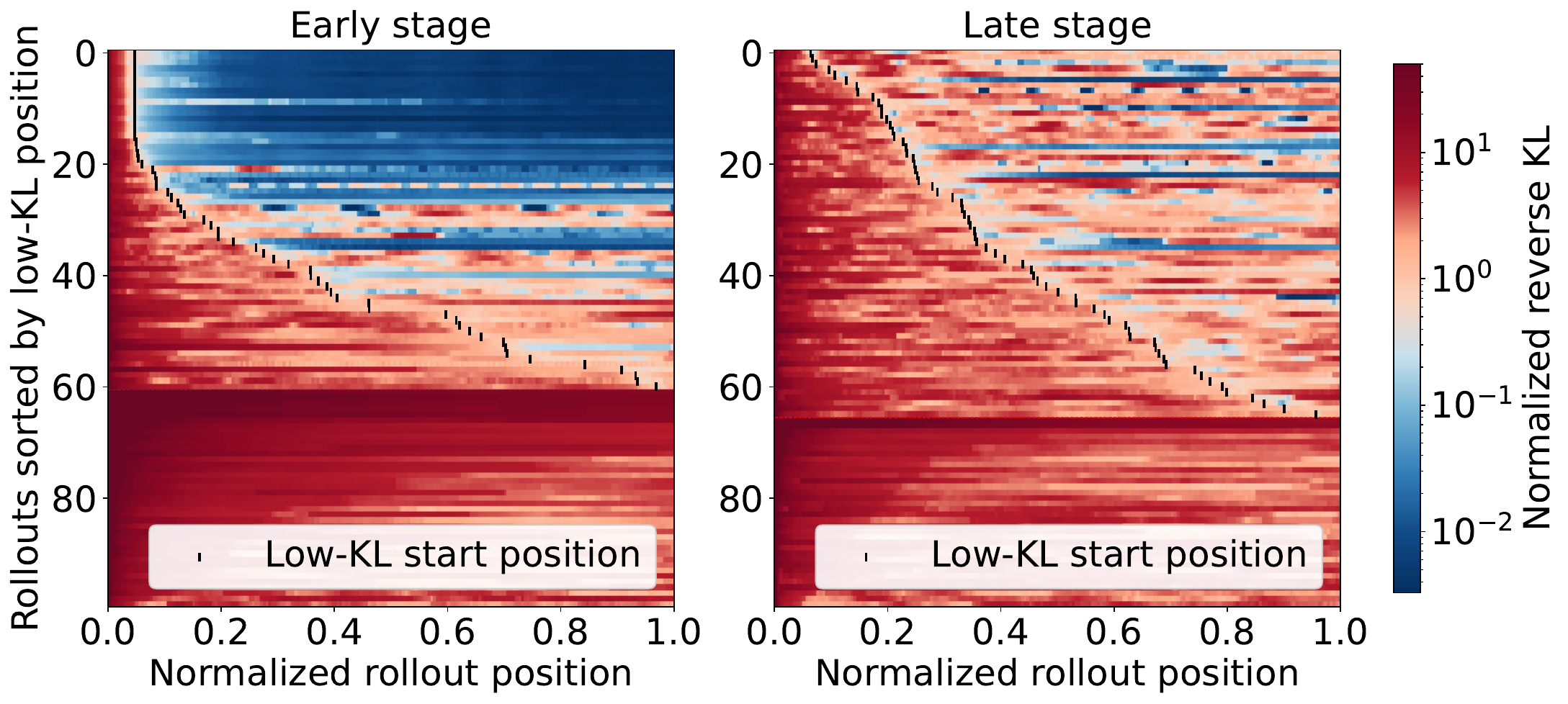}
        \caption{Per-token reverse KL heatmap.}
        \label{fig:kl-heatmap}
    \end{subfigure}
    \hspace{-2mm}
    \begin{subfigure}[c]{0.49\linewidth}
        \centering
        \includegraphics[width=\linewidth]{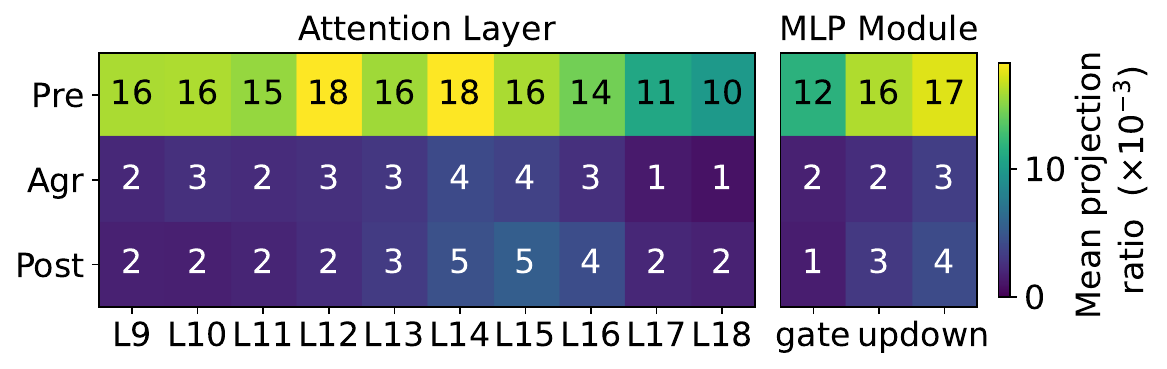}
        \caption{Principal-subspace projection ratio by phase.}
        \label{fig:subspace-projection}
    \end{subfigure}
    \hspace{-2mm}
    \begin{subfigure}[c]{0.95\linewidth}
        \centering
        \includegraphics[width=\linewidth]{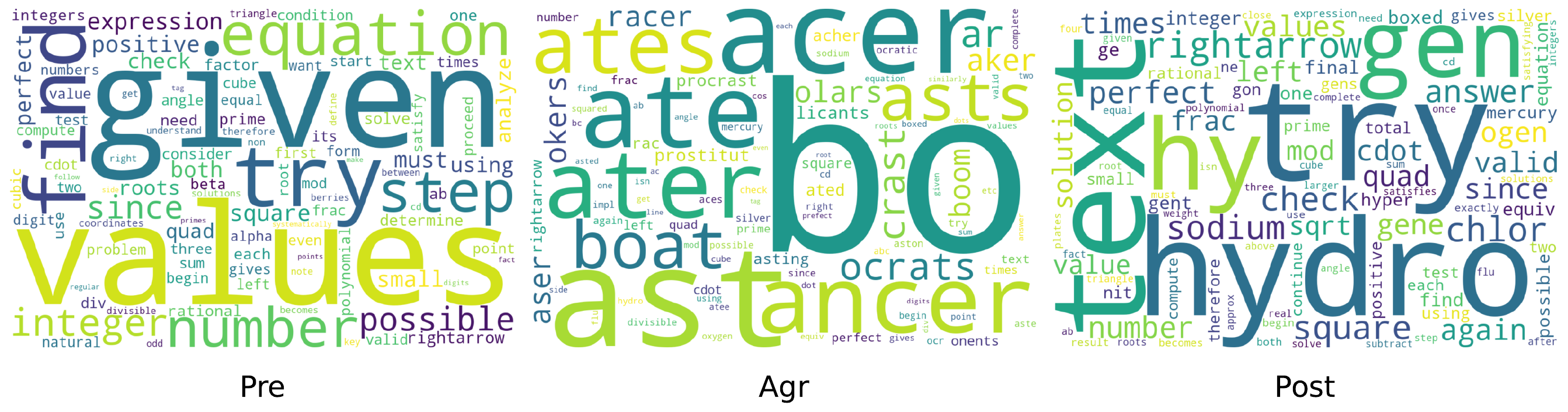}
        \caption{Teacher top-token word clouds by phase.}
        \label{fig:wordcloud}
    \end{subfigure}
    \caption{
    Evidence for low-KL agreement and weakened update alignment in OPD rollouts.
    \textbf{(a)} Per-token KL at early and late training stages, with rollouts sorted by the first low-KL position.
    Low-KL regions appear in many rollouts and occur earlier in early training.
    \textbf{(b)} OPD gradients in respective rollout stages are projected onto the principal update subspace.
    Pre-agreement gradients align much more strongly with the principal update directions than agreement and post-agreement gradients, suggesting that low-KL agreement marks a transition to less effective optimization signal.
    \textbf{(c)} Word clouds of the teacher's high-probability tokens in each phase: pre-agreement is dominated by reasoning-related vocabulary, while (post-)agreement shifts toward local fragments.
    }
    \vspace{-4mm}
    \label{fig:motivation-overview}
\end{figure*}

\section{Low-KL Agreement in OPD Rollouts}
\label{sec:motivation}

We motivate \SLOT{} by analyzing student rollouts during OPD training and identifying a recurring low-KL agreement pattern~(\S\ref{sec:mot-distribution}): at certain trajectory depth, the student and teacher often maintain sustained low reverse KL.
This pattern naturally partitions each rollout into three phases: \emph{pre-agreement}, \emph{agreement}, and \emph{post-agreement}.
We study these phases from an optimization perspective~(\S\ref{sec:mot-subspace}), using gradient projections to assess their contribution to effective learning, and further validate the teacher supervision through a token-level visualization~(\S\ref{sec:mot-tokens}).
Together, these analyses show that the agreement and post-agreement phases provide little useful supervision or parameter movement, motivating our proposed method.

\subsection{Low-KL Dynamics}
\label{sec:mot-distribution}

\paragraph{Setup.}
We sample $N = 100$ student rollouts at each of two training stages---\emph{early} (within the first 10\% of optimizer steps) and \emph{late} (near the end of training).
For each rollout, we record the per-token reverse KL $d_t$ and visualize it as a heatmap in \cref{fig:kl-heatmap}, with rollout position on the horizontal axis and rollout index on the vertical axis; color intensity encodes $\log d_t$.
Low-KL regions are marked using the calibrated threshold adopted by KAT.

\paragraph{Observations.}
The heatmap reveals three robust phenomena.
First, \textbf{low-KL regions are prevalent}: more than 60\% of rollouts exhibit at least one extended low-KL segment.
Second, \textbf{early-training rollouts enter low-KL earlier}: the median onset position of the first low-KL segment in early-stage rollouts is markedly smaller than in late-stage rollouts, indicating that an under-trained student more quickly drifts into agreements with the teacher.
Third, \textbf{low-KL regions tend to be followed by weaker KL signals}: after a rollout enters a low-KL segment, the subsequent reverse KL often remains lower than its pre-segment level rather than fully returning to earlier high-KL behavior.
This trend is especially pronounced in early training, where many rollouts exhibit extremely small KL values after entering such regions, indicating that the token-level OPD signal has largely lost update strength.
This observation suggests that low-KL regions may mark a qualitative change in the usefulness of later rollout tokens, motivating a closer analysis of the optimization signal they provide.

\subsection{Gradient Geometry}
\label{sec:mot-subspace}

The observations of \S\ref{sec:mot-distribution} induce a three-phase decomposition of each rollout:
\emph{pre-agreement} positions before the first extended low-KL segment, \emph{agreement} positions inside the segment, and \emph{post-agreement} positions after it.
We now ask whether the OPD updates from these phases align with the principal parameter directions during training.

\paragraph{Principal Update Subspace via Top-$k$ SVD.}
Let $\theta_0$ denote the parameters at the start of training and $\theta^\star$ that of the final analyzed checkpoint.
Inspired by ~\cite{cai2026learning}, for each layer, we form the net learned update $\Delta W = \theta^\star - \theta_0$ and compute its rank-$k$ truncated SVD~\cite{koren2009matrix},
\begin{equation}
\Delta W \;\approx\; U_k \Sigma_k V_k^\top .
\end{equation}
The top-$k$ singular directions define a {principal update subspace}: the dominant directions along which the model parameters changed over training.
Projection onto this subspace therefore provides a proxy for how much an update aligns with the main parameter movement learned by the model.

\paragraph{Projection Magnitude.}
At a selected analysis checkpoint $\theta_k$, and for each rollout phase $\phi \in \{\text{pre}, \text{agr}, \text{post}\}$, we form the phase-conditional gradient $g^{(\phi)}_k$ by accumulating per-token OPD gradients only at positions belonging to phase $\phi$ on a held-out probe set.
We then report the fraction of $g^{(\phi)}_k$'s squared norm that lies inside the layer's principal update subspace,
\begin{equation}
\rho^{(\phi)}_k \;=\; 
\frac{\| P_{U_k} g^{(\phi)}_k \|^2}
{\| g^{(\phi)}_k \|^2},
\end{equation}
where $P_{U_k} = U_k U_k^\top$ projects onto the principal update subspace.
A larger $\rho^{(\phi)}_k$ indicates that phase $\phi$ contributes gradient signal more aligned with the model's principal parameter update, while a smaller value indicates weaker alignment.

\paragraph{Observations.}
\Cref{fig:subspace-projection} shows a clear ordering across phases.
The \emph{pre-agreement} phase attains the largest projection ratio; the \emph{agreement} and \emph{post-agreement} phases are substantially smaller.
This comparison is especially informative for attention and final MLP parameters, which carry task-relevant updates in OPD training.
The result suggests that once a rollout enters a low-KL agreement region, the subsequent gradient signal becomes less aligned with the principal directions of parameter movement; moreover, the post-agreement segment does not fully recover this alignment even when the per-token KL increases again.
Thus, agreement and post-agreement tokens may provide less efficient supervision than pre-agreement tokens, motivating us to examine whether continuing to roll out through these regions is worthwhile.

\subsection{Teacher Vocabulary Shift}
\label{sec:mot-tokens}

To further understand what supervision the teacher provides in different phases, we visualize the tokens to which the teacher assigns high probability.

\paragraph{Setup.}
For each rollout position, we take the teacher's top-5 tokens under $\teacher(\cdot \given h_t)$, filter out special characters and non-content artifacts, and aggregate the remaining tokens separately over three phases.
We visualize the resulting empirical token distributions as word clouds in \cref{fig:wordcloud}.

\paragraph{Observations.}
The word clouds show a clear shift in the teacher's high-probability vocabulary across phases.
In the \emph{pre-agreement} phase, the teacher frequently assigns mass to reasoning-related tokens such as mathematical quantities, equation terms, and discourse markers, suggesting that the supervision still reflects useful intermediate reasoning.
In the \emph{agreement} phase, the distribution becomes more concentrated on local surface continuations and problem-specific fragments, with fewer tokens that clearly indicate stepwise reasoning.
The \emph{post-agreement} phase partially recovers content words, but its vocabulary remains noticeably different from the pre-agreement phase and contains more generic continuation tokens.
Together with the analysis in \S\ref{sec:mot-subspace}, this visualization suggests that low-KL regions are not merely low-disagreement regions; they also correspond to a qualitative change in the teacher's token-level supervision, motivating us to consider whether these later tokens are worth generating and training on.

\section{Method: Online Termination with Low-KL Agreement}
\label{sec:method}

Based on the analysis in \S\ref{sec:motivation}, we propose \SLOT{}, an online rollout termination method for OPD.
\SLOT{} uses the teacher--student reverse KL already computed during OPD to detect sustained low-KL agreement at sufficient rollout depth.
It calibrates a threshold from recent rollouts, monitors a sliding-window KL statistic during generation, and terminates the rollout once the statistic remains low for consecutive windows.
The truncated suffix is not generated and contributes no OPD supervision.

\subsection{Local KL Statistic}
\label{sec:slot-stat}

Consider a prompt $x$ and a student rollout $\hat{y}=(\hat{y}_1,\ldots,\hat{y}_T)$.
At token position $t$, let $h_t=(x,\hat{y}_{<t})$ denote the current prefix, and let
$p_t=\pi_\theta(\cdot \mid h_t)$ and $q_t=\pi_T(\cdot \mid h_t)$ denote the student and teacher next-token distributions, respectively.
We measure their local disagreement by the reverse KL $d_t$ in \cref{eq:kl}.
Directly thresholding $d_t$ at a single token is unreliable, since token-level KL can fluctuate due to punctuation, formatting tokens, or local lexical choices.
We therefore track a local average over a window of $W$ tokens:
\begin{equation}
\label{eq:zt}
z_t \;=\; \frac{1}{W} \sum_{i=t-W+1}^{t} d_i, \qquad t \ge W,
\end{equation}
where $W$ is the sliding-window size and $z_t$ summarizes the recent teacher--student disagreement around position $t$.
A small $z_t$ indicates that the teacher and student have remained close over a local region, which is the signal used by \SLOT{} to detect sustained low-KL agreement.

\subsection{Adaptive Threshold Calibration}
\label{sec:slot-threshold}

The scale of KL can vary substantially across training, so a fixed threshold is difficult to choose a priori.
\SLOT{} therefore calibrates the low-KL threshold from recent rollouts.
For each completed rollout $\hat{y}$, we record the smallest sliding-window KL value:
\begin{equation}
m(\hat y) \;=\; \min_{t \ge W} z_t .
\end{equation}
Intuitively, $m(\hat y)$ summarizes how low the rollout's local teacher--student disagreement becomes once it is eligible for early stopping.
During the first $K$ optimizer steps, \SLOT{} runs in a warmup mode: no rollout is terminated early, and each full rollout contributes its minimum score $m(\hat y)$ to a FIFO buffer $\mathcal{B}$.
The buffer stores the most recent $B$ minimum scores, providing a rolling estimate of the current low-KL scale under the evolving student policy.

After warmup, at the start of each step $k>K$, \SLOT{} sets the stopping threshold from the buffer:
\begin{equation}
\label{eq:threshold}
s_k \;=\; Q_{1-\eta}(\mathcal{B}),
\end{equation}
where $Q_{1-\eta}$ denotes the $(1-\eta)$-quantile and $\eta$ controls the aggressiveness of stopping.
Smaller $\eta$ yields a more conservative rule: for example, with $\eta=10\%$, the threshold is the 90\% quantile of recent rollout minima, so a rollout is considered low-KL when its local window score falls below this more permissive threshold.
This calibrated threshold makes the rule data-driven and adaptive to the changing KL scale during optimization, avoiding hand-tuning an absolute KL cutoff.

\subsection{Online Trigger and Termination}
\label{sec:slot-trigger}

After warmup, \SLOT{} applies the calibrated threshold online during rollout.
Let $k$ denote the current optimizer step, $K$ the number of warmup steps, $z_t$ the sliding-window KL statistic ending at position $t$, and $s_k$ the threshold calibrated from the buffer.
We also introduce an exemption length $L_0$, which excludes the initial KL region from the stopping decision.
A rollout becomes eligible for termination only when $k>K$ and $t \ge L_0 + W$.
This exemption is important because early in a rollout, reverse KL often drops rapidly from a high-disagreement regime to a lower one while the teacher may still provide meaningful corrective signal.
\SLOT{} therefore focuses on sustained low-KL agreement that appears after this early transition.

For each eligible rollout, we maintain a counter $c$ for consecutive low-KL windows.
The counter is initialized to zero for each rollout and updated as
\begin{equation}
\label{eq:counter}
c \leftarrow 
\begin{cases}
c + 1, & z_t < s_k, \\
0, & z_t \ge s_k .
\end{cases}
\end{equation}
Here, $z_t < s_k$ means that the window ending at position $t$ is classified as low-KL.
We define $T$ as the consecutive trigger length: the number of consecutive low-KL windows required before termination is confirmed.
This requirement prevents the rule from firing on isolated KL dips and makes termination depend on a sustained low-KL pattern.

When $c \ge T$ at position $t$, the trigger is confirmed.
Since this confirmation occurs only after observing $T$ consecutive low-KL windows, the effective truncation point is set by backtracking to the start of the first low-KL window in this consecutive run.
No further student tokens are generated for that prompt, and tokens after the effective truncation point do not contribute OPD updates.

\paragraph{Remarks}
\label{sec:slot-remarks}
We defer the full \SLOT{} pseudocode and the streaming teacher-scoring implementation to \cref{sec:appendix-alg}.
The added online state is $O(1)$ per token, and the teacher signal already required by OPD is reused without additional rollouts.

\section{Experiments}
\label{sec:experiments}

\begin{table*}[t]
\centering
\small
\setlength{\tabcolsep}{2.5pt}
\begin{tabular}{lcccccccc|cc|c}
\toprule
& \multicolumn{2}{c}{AMC} & \multicolumn{2}{c}{MATH500} & \multicolumn{2}{c}{Minerva} & \multicolumn{2}{c}{AIME24} & \multicolumn{2}{|c|}{Avg} & \multirow{2}{*}[-0.6ex]{\begin{tabular}[c]{@{}c@{}}Rollout\\Length\end{tabular}} \\
\cmidrule(lr){2-3} \cmidrule(lr){4-5} \cmidrule(lr){6-7} \cmidrule(lr){8-9} \cmidrule(lr){10-11}
Method & avg@$k$ & pass@$k$ & avg@$k$ & pass@$k$ & avg@$k$ & pass@$k$ & avg@$k$ & pass@$k$ & avg@$k$ & pass@$k$ & \\
\midrule
Teacher & 58.28 & 77.11 & 83.18 & 94.20 & 29.23 & 41.18 & 27.92 & 63.33 & 49.65 & 68.96 & - \\
\midrule
\multicolumn{12}{l}{\textbf{\textit{Qwen3-1.7B-Base}}} \\
\midrule
Student & 10.09 & 40.96 & 20.55 & 65.60 & 5.38 & 24.27 & 1.67 & 13.33 & 9.42 & 36.04 & - \\
\midrule
OPD & 20.33 & 53.01 & 41.55 & 80.00 & 10.71 & 30.52 & 3.33 & 23.33 & 18.98 & 46.72 & 1013 \\
~ - \textit{Random Termination} & 19.88 & 46.99 & \textbf{44.50} & 79.20 & 10.98 & 28.68 & 3.54 & \textbf{33.33} & \underline{19.73} & 47.05 & 682 \\
~ - \textit{Fixed Prefix} & 20.63 & 49.40 & 40.78 & 78.00 & 10.39 & 33.46 & 4.27 & 30.00 & 19.02 & \underline{47.72} & 730 \\
\midrule
\textbf{\SLOT{}-OPD} & \textbf{21.24} & \textbf{56.63} & \underline{43.33} & \textbf{81.40} & \textbf{11.49} & \textbf{33.82} & \textbf{4.48} & \underline{30.00} & \textbf{20.14} & \textbf{50.46} & \textbf{456} \\
\midrule
\midrule
\multicolumn{12}{l}{\textbf{\textit{Qwen3-4B-Base}}} \\
\midrule
Student & 14.91 & 50.60 & 21.90 & 74.80 & 6.34 & 30.15 & 5.21 & 40.00 & 12.09 & 48.89 & - \\
\midrule
OPD & 45.48 & 73.49 & \underline{78.80} & 91.20 & 25.60 & 43.02 & 16.35 & \underline{43.33} & \underline{41.56} & 62.76 & 1947 \\
~ - \textit{Random Termination} & 44.13 & 68.68 & \textbf{79.00} & 91.40 & 24.95 & 45.59 & 14.38 & 33.33 & 40.62 & 59.75 &{863} \\
~ - \textit{Fixed Prefix} & 45.19 & 72.31 & 78.05 & 90.60 & 25.41 & 44.49 & 15.88 & \textbf{46.67} & 41.13 & \textbf{63.52} & 987 \\
\midrule
\textbf{\SLOT{}-OPD} & \textbf{46.69} & \textbf{75.90} & {78.23} & \textbf{91.80} & \textbf{26.54} & \textbf{46.75} & \textbf{16.56} & {36.67} & \textbf{42.01} & \underline{62.78} & \textbf{736} \\
\bottomrule
\end{tabular}
\caption{avg@$k$ and pass@$k$ accuracy on four benchmarks for two teacher--student pairs. We compare \SLOT{}-OPD against the zero-shot Student/Teacher endpoints, standard OPD, and two length-reduction baselines: {Random Termination} and {Fixed Prefix}. The rightmost column reports the average rollout length in tokens. Best per-column results in \textbf{bold}, second-best \underline{underlined}.}
\label{tab:main}
\end{table*}

\subsection{Setup}
\label{sec:exp-setup}

\paragraph{Models.}
We evaluate two teacher--student pairs.
The teacher is Qwen3-8B, and the students are Qwen3-0.7B and Qwen3-1.6B, respectively.
For each pair, the student is trained with OPD under the same training recipe. The implementation details are described in \cref{sec:impl}.

\paragraph{Benchmarks and Metrics.}
We evaluate mathematical reasoning performance on AMC, MATH500, MinervaMath, and AIME24.
We report avg@$k$ and pass@$k$ accuracy on each benchmark, as well as the average across benchmarks.
Following common evaluation budgets, we use $k=32$ for AIME24, and $k=8$ for the remaining benchmarks.


\paragraph{Baselines.}
We compare \SLOT{}-OPD against the following baselines.
\textbf{Teacher} and \textbf{Student} report the zero-shot performance of the original teacher and student models, respectively.
\textbf{OPD} is the standard on-policy distillation baseline.
We also compare against two OPD variants that reduce rollout length without using the proposed low-KL agreement signal.
\textbf{Random Termination} stops each rollout at a randomly sampled position, controlling for the effect of shortening rollouts without selecting specific low-KL regions.
\textbf{Fixed Prefix} caps the maximum response length at 1024, providing a simple length-based truncation baseline.
\textbf{\SLOT{}-OPD} is our full method, which terminates rollouts online when sustained low-KL agreement is detected.

\begin{figure}[t]
    \centering
    \begin{subfigure}[c]{0.495\linewidth}
        \centering
        \includegraphics[width=\linewidth]{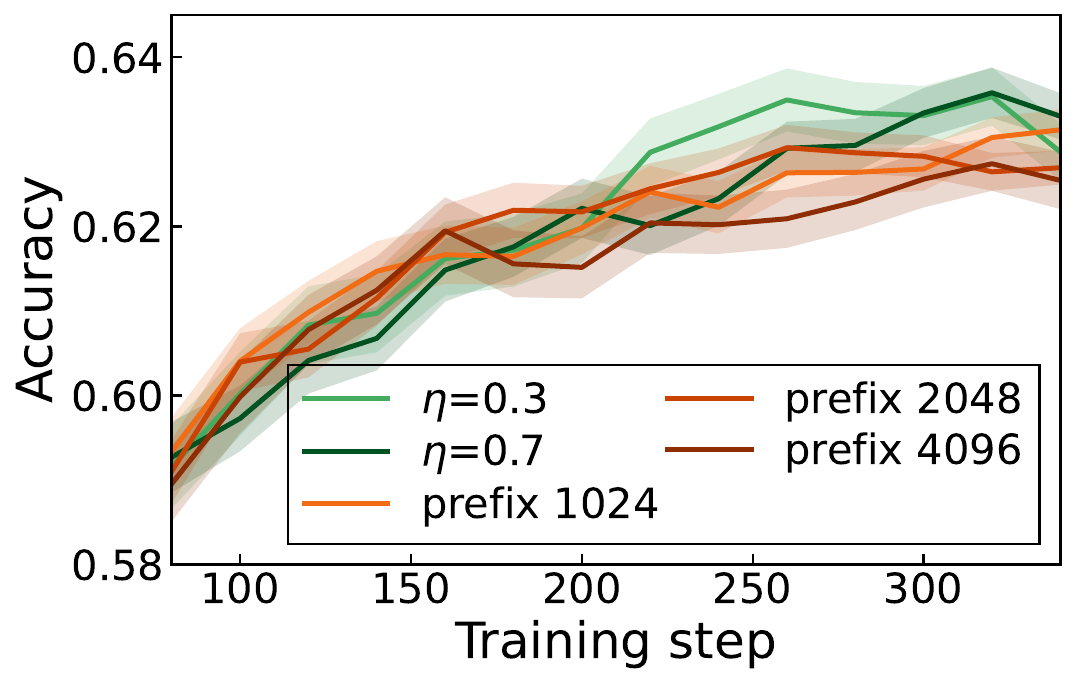}
        \label{fig:training-dynamics-acc}
    \end{subfigure}
    \hspace{-2mm}
    \begin{subfigure}[c]{0.495\linewidth}
        \centering
        \includegraphics[width=\linewidth]{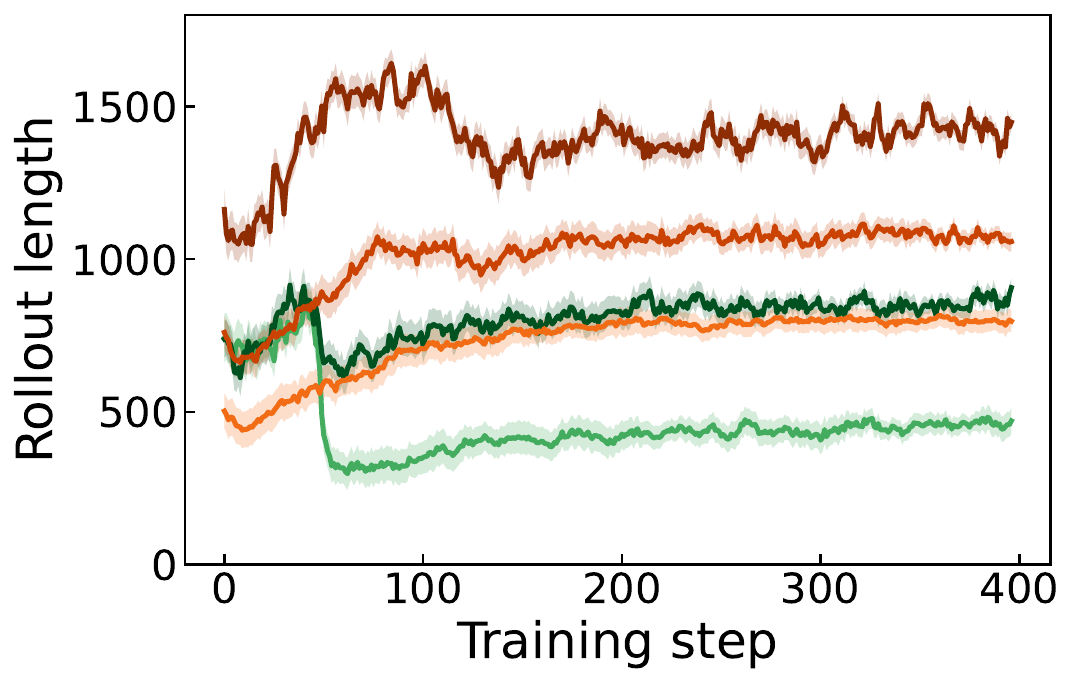}
        \label{fig:training-dynamics-len}
    \end{subfigure}
    \vspace{-5mm}
    \caption{Training dynamics of \SLOT{} versus fixed-prefix OPD with varying lengths. \SLOT{} reaches higher accuracy earlier while maintaining substantially shorter effective rollout lengths, indicating that adaptive termination of agreement traps improves the accuracy--compute trade-off over uniform prefix truncation.}
    \vspace{-3mm}
    \label{fig:training-dynamics}
\end{figure}

\subsection{Main Results}
\label{sec:exp-main}

\Cref{tab:main} reports avg@$k$ and pass@$k$ accuracy across benchmarks.
Overall, \SLOT{}-OPD improves both accuracy and rollout efficiency over standard OPD.
Averaged over the two student settings, \SLOT{} increases avg@$k$ from 30.27 to 31.08, a relative improvement of 2.66\%, and improves pass@$k$ from 54.74 to 56.62, a relative gain of 3.43\%.
Meanwhile, the average rollout length is reduced from 1480.0 to 596.0, saving 59.73\% of rollout tokens.

The comparison with length-reduction baselines further highlights the advantage of \SLOT{}.
Standard OPD keeps the full rollout and can therefore train on late tokens where the teacher signal has become weak.
Fixed Prefix truncates every rollout at a fixed length, ignoring the quality and importance of the current trajectory state.
Random Termination shortens rollouts at random positions and cannot distinguish useful reasoning prefixes from low-quality suffixes.
In contrast, \SLOT{} makes prefix-dependent online decisions based on sustained low-KL agreement between the student and teacher.
This allows it to preserve useful early supervision while terminating rollouts once the teacher signal becomes less informative, explaining why it achieves higher accuracy with substantially fewer rollout tokens.

\subsection{Analysis}
\label{sec:exp-analysis}
\label{sec:exp-pareto}
\label{sec:exp-ablations}

\paragraph{Training Dynamics Analysis.}
\Cref{fig:training-dynamics} compares \SLOT{} with fixed-prefix OPD variants of different prefix lengths, reporting validation accuracy and effective rollout length.
For both quantile settings, $\eta=0.3$ and $\eta=0.7$, \SLOT{} reaches the converged accuracy region earlier and achieves higher final accuracy, while maintaining substantially shorter effective rollouts.
This supports our central motivation: useful OPD supervision is concentrated in informative prefixes, so adaptive termination can improve the accuracy--compute trade-off without preserving uniformly long rollouts.

We also observe that longer fixed prefixes do not monotonically improve training.
In particular, the 4096-token prefix underperforms the 2048-token prefix despite exposing the student to more teacher supervision.
This suggests that late rollout tokens may provide weak or even harmful signals: once the student has drifted into an unrecoverable reasoning state, the teacher can no longer offer effective corrective guidance.
Thus, simply extending rollout length may dilute useful early-token supervision with low-quality suffix supervision.

\begin{figure}[t]
    \centering
    \includegraphics[width=\linewidth]{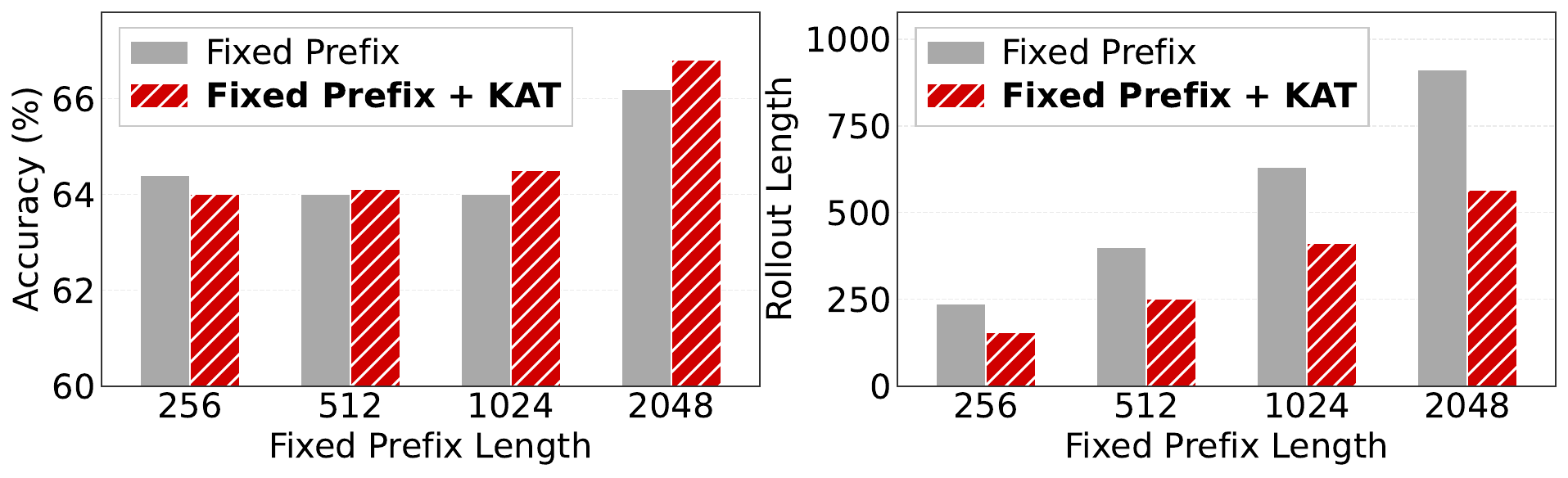}
    \caption{Applying \SLOT{} on top of fixed-prefix OPD across prefix lengths
$\{256, 512, 1024, 2048\}$ (Qwen3-1.7B-Base). \textbf{Left}: accuracy;
\textbf{Right}: effective rollout length. At moderate-to-large budgets, \SLOT{}
(hatched) improves accuracy over the fixed-prefix baseline (solid) while shortening
rollouts, with the length gap widening as the budget grows.}
    \label{fig:fixed_kat}
    \vspace{-4mm}
\end{figure}

\paragraph{\SLOT{} on Top of Different Prefix Lengths.}
To disentangle \SLOT{} from the choice of prefix budget, we apply it on top of fixed-prefix OPD across prefix lengths $\{256, 512, 1024, 2048\}$ (Fig.~\ref{fig:fixed_kat}).
At moderate-to-large budgets, \SLOT{} improves accuracy while shortening the effective rollout, as it removes uninformative suffix tokens and concentrates supervision on informative prefixes; the length gap also widens with larger budgets, which admit more post-trap suffix to truncate.
At the smallest budget ($256$), however, \SLOT{} yields no accuracy gain and even a slight drop, since the short cap already discards most of the low-KL agreement region, and further truncation begins to remove useful early signal as well.
This indicates that \SLOT{} is complementary to length-based truncation rather than a substitute: a fixed prefix bounds the worst-case length, while \SLOT{} adaptively removes the low-KL agreement region within that budget, improving the accuracy--compute trade-off whenever the budget is large enough to contain such a region.

\paragraph{Training Efficiency.}
\Cref{tab:efficiency} reports the end-to-end training cost of \SLOT{}-OPD compared with standard OPD and the Fixed-Prefix baseline.
\SLOT{}-OPD consistently reduces both wall-clock time and compute across model scales.
On Qwen3-1.7B-Base, it reduces training time from $12.30$h to $5.20$h ($-57.7\%$) and compute from $1.98$ to $1.30$ EFLOPs ($-34.3\%$) relative to OPD.
On Qwen3-4B-Base, it trains $2.4\times$ faster ($11.00$h vs.\ $26.28$h) and uses $59.4\%$ fewer EFLOPs ($2.30$ vs.\ $5.67$).
Compared with Fixed Prefix, \SLOT{}-OPD remains $23$--$35\%$ faster and uses $16$--$24\%$ fewer FLOPs, showing that adaptive termination is more efficient than a uniform length cap.
Together with the accuracy gains in \Cref{tab:main}, these results show that \SLOT{}-OPD achieves a better accuracy--compute trade-off.

\begin{figure*}[t]
    \centering
    \begin{subfigure}[c]{0.33\linewidth}
        \centering
        \includegraphics[width=\linewidth]{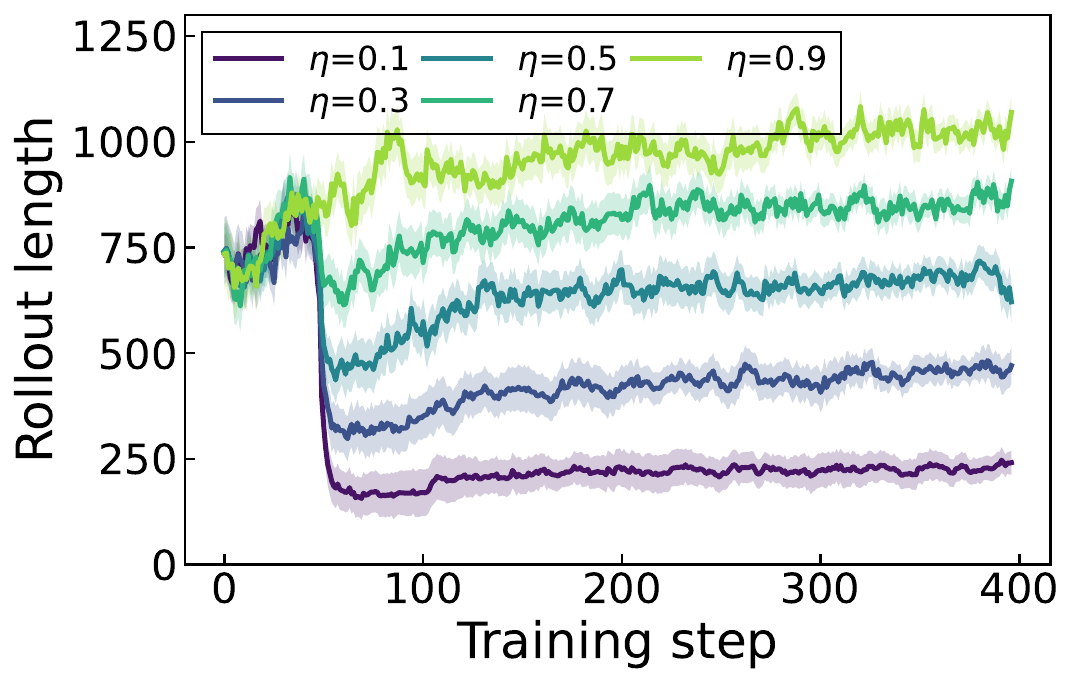}
        \caption{Effective rollout length vs.\ $\eta$.}
        \label{fig:eta-length}
    \end{subfigure}
    \hspace{-2mm}
    \begin{subfigure}[c]{0.33\linewidth}
        \centering
        \includegraphics[width=\linewidth]{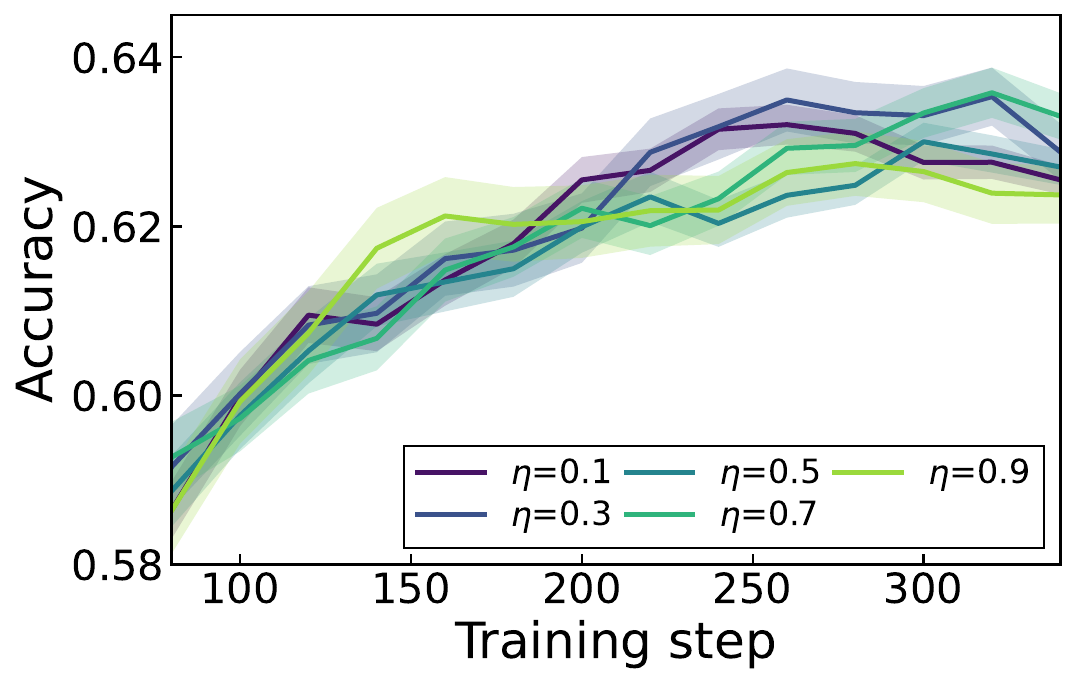}
        \caption{Average accuracy vs.\ $\eta$.}
        \label{fig:eta-accuracy}
    \end{subfigure}
    \begin{subfigure}[c]{0.33\linewidth}
        \centering
        \includegraphics[width=\linewidth]{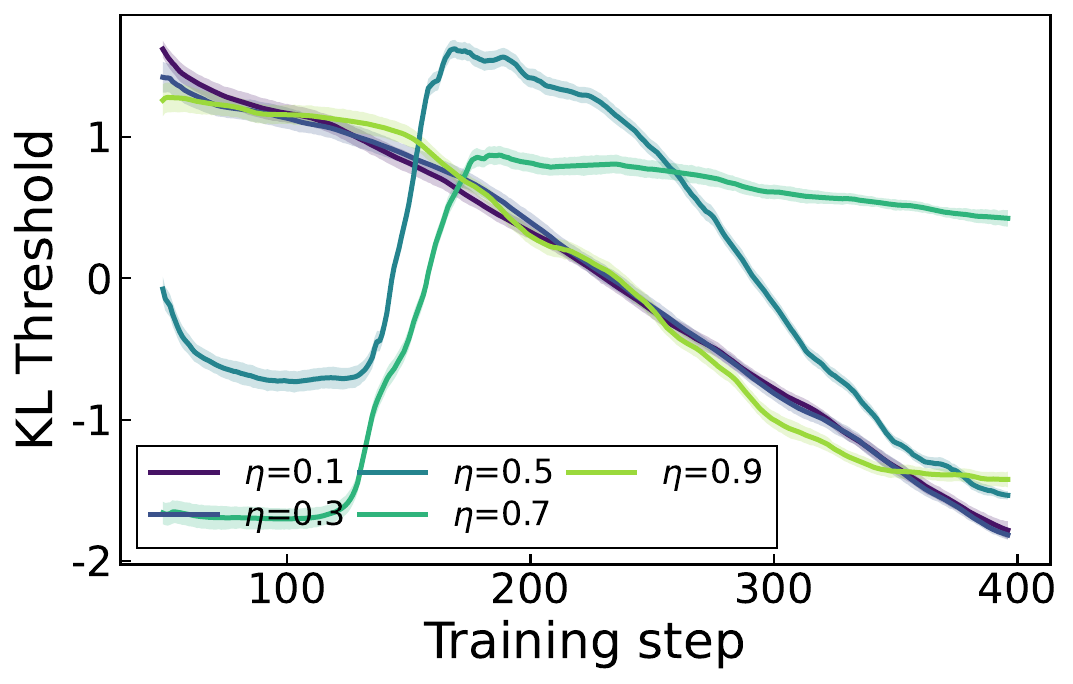}
        \caption{Normalized adaptive threshold.}
        \label{fig:eta-threshold}
    \end{subfigure}
    \vspace{-2mm}
    \caption{Effect of the quantile parameter $\eta$ on \SLOT{}. Smaller $\eta$ produces a more permissive threshold and therefore shorter effective rollouts (\subref{fig:eta-length}), which generally yields higher average accuracy (\subref{fig:eta-accuracy}), supporting the view that many late rollout tokens contribute weak supervision. The normalized adaptive threshold (\subref{fig:eta-threshold}) decreases as training proceeds, with a transient plateau near step 100 that lets \SLOT{} filter noisy suffixes during unstable phases.}
    \vspace{-2mm}
    \label{fig:eta-analysis}
\end{figure*}

\begin{table}[t]
\centering
\small
\setlength{\tabcolsep}{10pt}
\begin{tabular}{lcc}
\toprule
Method & GPU Hours & EFLOPs \\
\midrule
\multicolumn{3}{l}{\textit{Qwen3-1.7B-Base}} \\
\midrule
OPD & 12.30 & 1.98 \\
~ - \textit{Fixed Prefix} & 7.96 & 1.54 \\
\textbf{\SLOT{}-OPD} & \textbf{5.20} & \textbf{1.30} \\
\midrule
\multicolumn{3}{l}{\textit{Qwen3-4B-Base}} \\
\midrule
OPD & 26.28 & 5.67 \\
~ - \textit{Fixed Prefix} & 14.28 & 3.04 \\
\textbf{\SLOT{}-OPD} & \textbf{11.00} & \textbf{2.30} \\
\bottomrule
\end{tabular}
\caption{End-to-end training cost in GPU hours and EFLOPs. \SLOT{}-OPD reduces training time by up to $2.4\times$ and compute by up to $59\%$ over standard OPD, while also outperforming the fixed-prefix baseline.}
\label{tab:efficiency}
\end{table}

\begin{figure}[t]
    \centering
    \includegraphics[width=\linewidth]{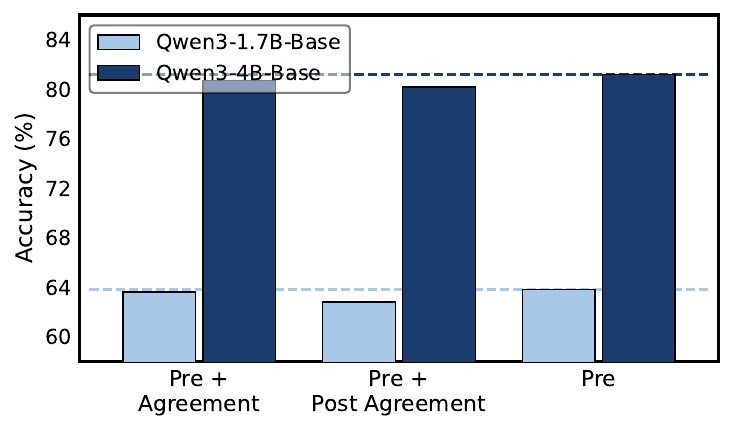}
    \caption{Stage-wise ablation on MATH500. Each rollout is split into pre-agreement, agreement, and post-agreement stages; dashed lines denote \SLOT{}-OPD, which trains only on pre-agreement tokens. Adding back any truncated stage reduces accuracy, with post-agreement causing a larger drop, indicating that supervision further deteriorates after the agreement trap.}
    \label{fig:stage-ablation}
    \vspace{-4mm}
\end{figure}

\paragraph{Stage-wise Supervision Ablation.}
The low-KL agreement trap partitions each rollout into pre-agreement, agreement, and post-agreement stages.
To test this decomposition, we retrain OPD by adding different truncated stages back to the \SLOT{} training signal.
\Cref{fig:stage-ablation} shows that adding either the agreement or post-agreement stage reduces accuracy compared with \SLOT{}-OPD, and adding the post-agreement stage causes a larger drop.
This indicates that supervision after entering the agreement trap is already weak, and the later post-trap suffix provides even lower-quality signals.
These results support \SLOT{}'s design choice of terminating rollouts at the trap entry and training only on the informative pre-agreement prefix.

\paragraph{Impact of Quantile $\eta$.}
\Cref{fig:eta-analysis} studies how the quantile parameter $\eta$ affects rollout length, accuracy, and the adaptive threshold.
Under our threshold convention, smaller $\eta$ gives a more permissive threshold, making low-KL windows easier to truncate and thus producing shorter effective rollouts (\cref{fig:eta-length}).
The accuracy curves in \cref{fig:eta-accuracy} show that smaller $\eta$ often performs better, suggesting that more aggressive removal of low-quality suffixes can improve the effective training signal.
This further supports our main observation that many late rollout tokens provide limited useful supervision.
Finally, \cref{fig:eta-threshold} shows that the normalized adaptive threshold generally decreases during training as the student becomes closer to the teacher.
Around step 100, several settings exhibit a temporary plateau or increase, likely due to training instability enlarging the estimate.
This larger threshold can be beneficial in unstable phases by filtering more noisy suffixes, before relaxing as training stabilizes.

\section{Related Work}
\label{sec:related}

\paragraph{On-policy Distillation.}
On-policy distillation (OPD) trains students on self-generated states with dense teacher supervision, reducing off-policy train--test mismatch while improving sample efficiency over sparse reward RL \citep{agarwal2024policy}.
Recent work studies its mechanisms and limitations, including early alignment with the eventual update trajectory \citep{cai2026learning} and factors such as teacher--student compatibility, shared high-probability tokens, and response-length sweet spots \citep{li2026rethinking}.
Other methods refine the OPD objective or teacher signal with adaptive targets, entropy-aware KL, relaxed imitation, or generalized KL constraints \citep{jang2026stable,jin2026entropy,ko2026scaling,yang2026learning}.
Efficiency-oriented work such as prefix-only OPD truncates rollouts based on the observation that useful supervision often concentrates in early reasoning prefixes \citep{zhang2026fast}.
Complementarily, we identify persistent low-KL agreement on corrupted prefixes as a trajectory-dependent failure mode and use it as an online signal to terminate uninformative suffixes without changing the OPD loss.

\paragraph{Rollout Filtering in RL.}
Recent RL post-training methods filter generated rollouts because many are redundant, noisy, or harmful.
Existing approaches operate at different granularities, including rollout-level sampling or overlong filtering \citep{yu2026dapo,xu2025not}, prompt/task-level filtering by learning utility or difficulty \citep{zheng2026act,bae2026online}, and response-level truncation or confidence-based trace filtering \citep{fan2025truncated,fu2025deep}.
They typically rely on rewards, difficulty, length, or confidence.
Complementarily, we filter within OPD trajectories using the teacher--student KL itself, terminating suffixes where persistent low KL indicates weak corrective supervision rather than useful agreement.

\section{Conclusion}
We identified low-KL agreement traps in on-policy distillation, a failure mode where the teacher and student remain locally close on degraded rollout prefixes but provide weak corrective supervision. To address this, we proposed KAT, an online termination rule that reuses the reverse KL already computed in OPD to detect sustained low-KL agreement with a training-adaptive threshold. By truncating uninformative suffixes without changing the OPD objective or adding auxiliary models, KAT improves the performance while substantially reducing rollout length, showing that KAT can make OPD both more effective and efficient.

\section*{Limitations}
This work focuses on mathematical reasoning tasks. Although the proposed termination rule is simple and model-agnostic, its behavior on broader domains such as code generation, open-ended instruction following, and multi-turn interaction remains to be further studied. In addition, KAT introduces several hyperparameters. While these parameters are intuitive and inexpensive to tune, more systematic guidance could further improve usability across training settings.

\bibliography{custom}

\newpage
\appendix

\section*{Appendix}

\section{Analysis on Qwen3-4B-Base Student}
\label{sec:appendix-4b}

\begin{figure}[h]
    \centering
    \begin{subfigure}[c]{0.495\linewidth}
        \centering
        \includegraphics[width=\linewidth]{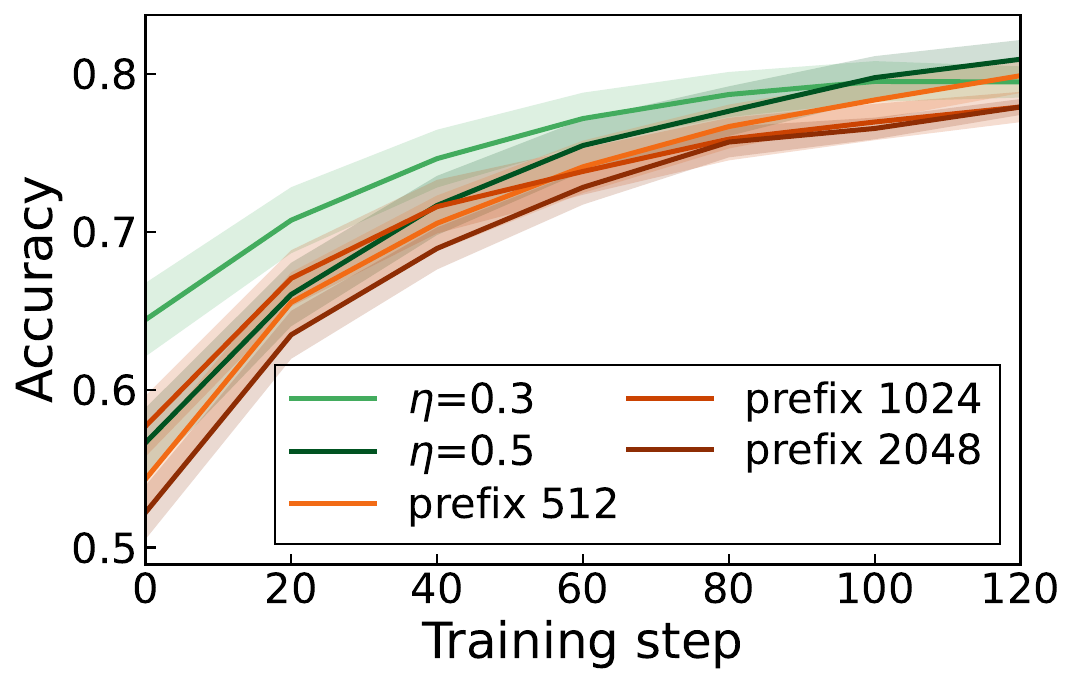}
        \caption{Accuracy over optimizer steps.}
        \label{fig:training-dynamics-acc-4b}
    \end{subfigure}
    \hspace{-1mm}
    \begin{subfigure}[c]{0.495\linewidth}
        \centering
        \includegraphics[width=\linewidth]{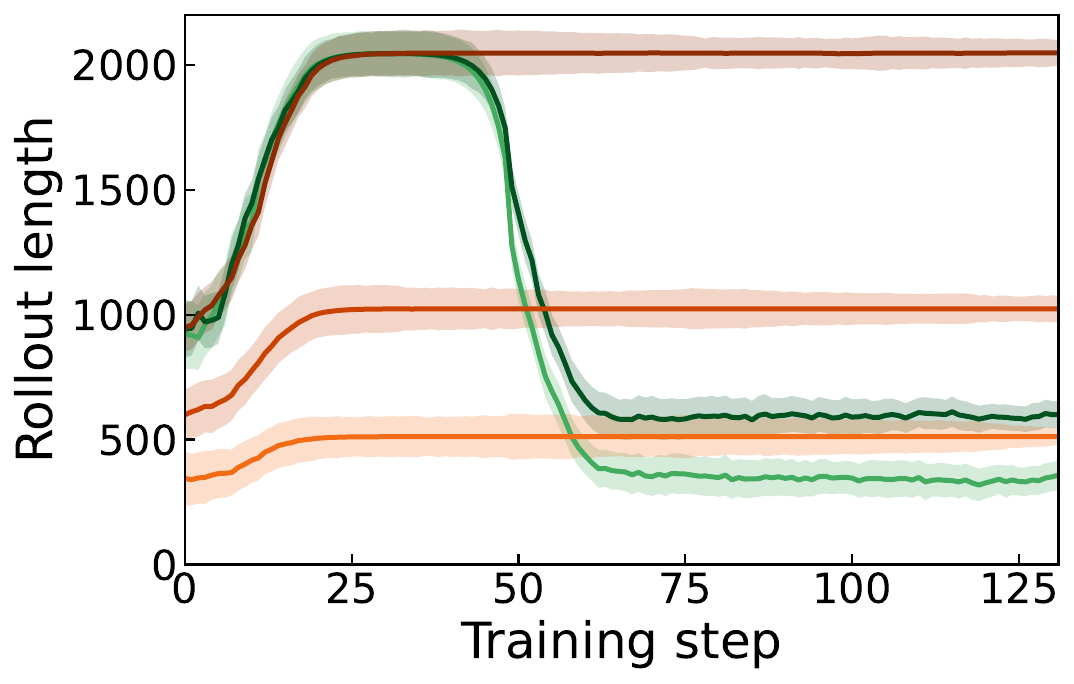}
        \caption{Effective rollout length over optimizer steps.}
        \label{fig:training-dynamics-len-4b}
    \end{subfigure}
    \caption{Training dynamics of \SLOT{} versus fixed-prefix OPD with varying lengths on the Qwen3-4B-Base student. As in the smaller-student setting, \SLOT{} reaches the converged accuracy region earlier (\subref{fig:training-dynamics-acc-4b}) while maintaining substantially shorter effective rollout lengths (\subref{fig:training-dynamics-len-4b}).}
    \label{fig:training-dynamics-4b}
\end{figure}

\begin{figure*}[h]
    \centering
    \begin{subfigure}[c]{0.33\linewidth}
        \centering
        \includegraphics[width=\linewidth]{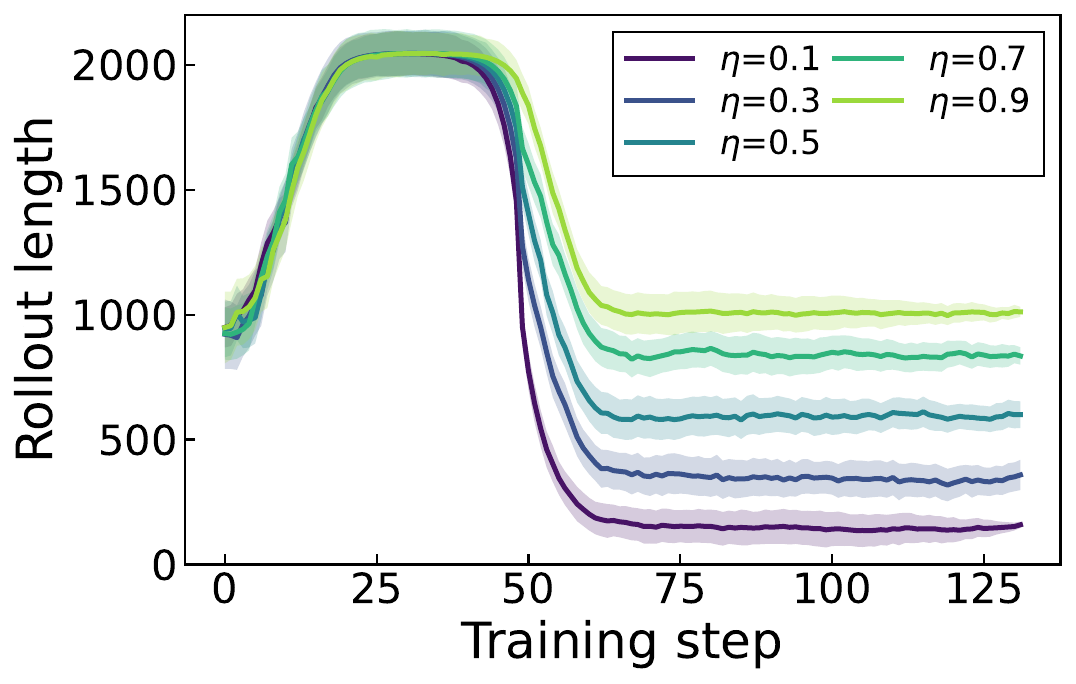}
        \caption{Effective rollout length vs.\ $\eta$.}
        \label{fig:eta-length-4b}
    \end{subfigure}
    \hspace{-2mm}
    \begin{subfigure}[c]{0.33\linewidth}
        \centering
        \includegraphics[width=\linewidth]{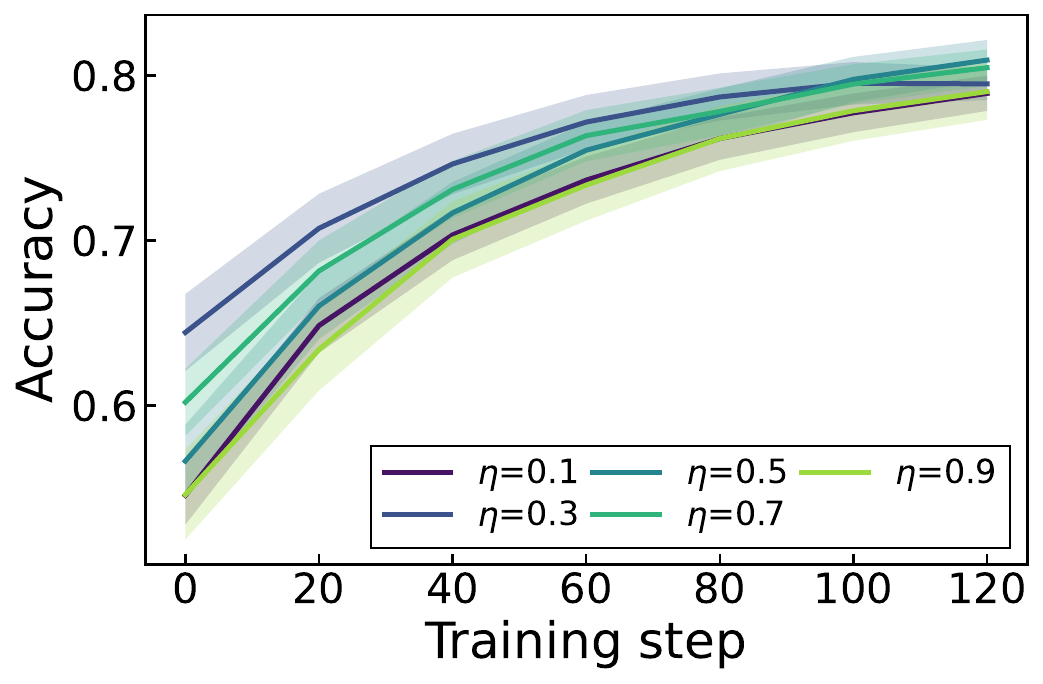}
        \caption{Average accuracy vs.\ $\eta$.}
        \label{fig:eta-accuracy-4b}
    \end{subfigure}
    \begin{subfigure}[c]{0.33\linewidth}
        \centering
        \includegraphics[width=\linewidth]{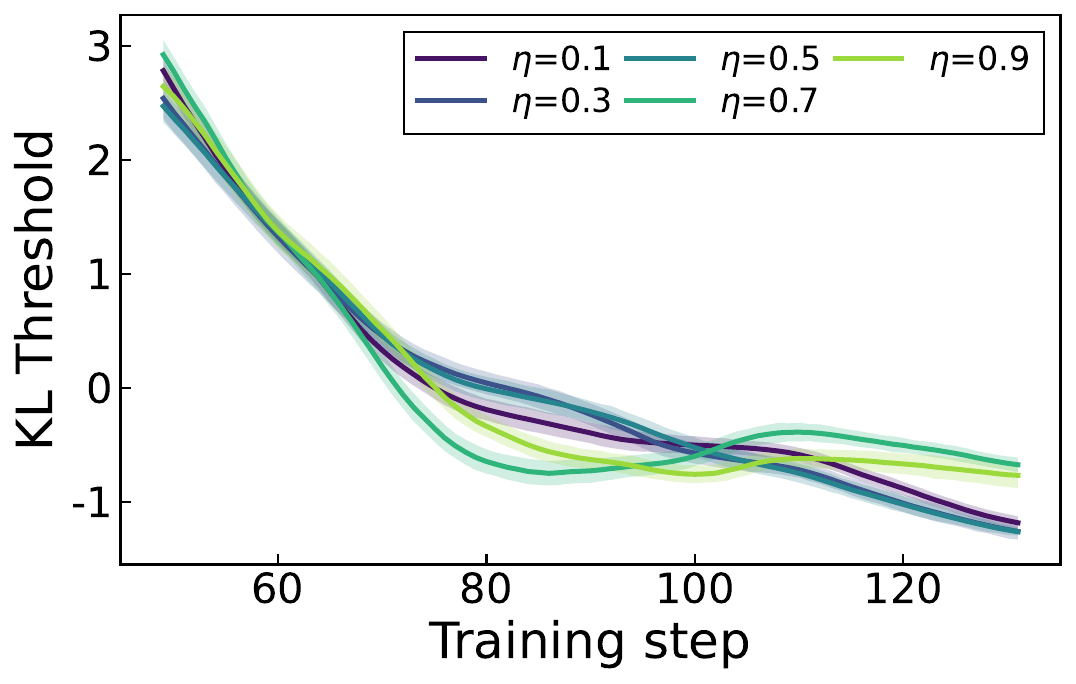}
        \caption{Normalized adaptive threshold.}
        \label{fig:eta-threshold-4b}
    \end{subfigure}
    \caption{Effect of the quantile parameter $\eta$ on \SLOT{} for the Qwen3-4B-Base student. Smaller $\eta$ again yields shorter effective rollouts (\subref{fig:eta-length-4b}) and generally higher average accuracy (\subref{fig:eta-accuracy-4b}), and the normalized adaptive threshold (\subref{fig:eta-threshold-4b}) decreases as training proceeds.}
    \label{fig:eta-analysis-4b}
\end{figure*}

We report the analysis on the larger Qwen3-4B-Base student.
\Cref{fig:training-dynamics-4b} reports the training dynamics of \SLOT{} against fixed-prefix OPD variants, and \cref{fig:eta-analysis-4b} reports the effect of the quantile parameter $\eta$ on rollout length, accuracy, and the normalized adaptive threshold.

\paragraph{Training dynamics.}
\Cref{fig:training-dynamics-4b} mirrors the trend observed for the smaller student.
\SLOT{} converges to its accuracy plateau noticeably earlier than the fixed-prefix variants while operating with a substantially shorter effective rollout length throughout training.
The relative ordering of fixed-prefix variants is also consistent: blindly increasing the prefix length does not monotonically improve accuracy, again indicating that uniformly long rollouts contain late tokens whose teacher signal has become weak or even harmful.
This reinforces that the gains of \SLOT{} are not a quirk of the smaller student but arise from selectively removing low-quality suffix supervision.

\paragraph{Quantile sensitivity.}
\Cref{fig:eta-analysis-4b} shows that the qualitative effect of $\eta$ is preserved at 4B scale.
More permissive $\eta$ shortens rollouts (\cref{fig:eta-length-4b}) and tends to produce equal or higher average accuracy (\cref{fig:eta-accuracy-4b}), confirming that aggressive removal of low-quality suffix tokens remains beneficial as the student grows.
The normalized adaptive threshold in \cref{fig:eta-threshold-4b} again drifts downward as the student becomes closer to the teacher, with a similar transient bump in the early training phase that lets \SLOT{} filter noisier suffixes while training is still unstable.
Together with the 1.7B results, these plots indicate that the behavior of \SLOT{} transfers consistently across student scales.

\section{Algorithm and Online Teacher Scoring}
\label{sec:appendix-alg}

\paragraph{Algorithm.}
The full \SLOT{}-OPD procedure described in \cref{sec:method} is summarized as pseudocode in \cref{alg:slot}.
The only additional online state is the rolling sum used to compute the sliding-window statistic $z_t$ and the consecutive-window counter $c$.
Both can be updated incrementally during generation, so the bookkeeping cost is $O(1)$ per token.

\paragraph{Online teacher scoring.}
Online termination requires the teacher to score student-generated prefixes during rollout, rather than after the full sequence has been generated.
We implement this in a streaming manner: as the student generates new tokens, the teacher receives the same growing prefix and produces the corresponding next-token distribution used to compute the reverse KL.
To keep this efficient, the teacher also maintains and reuses its KV cache, so each new scoring step only incurs incremental computation instead of recomputing the entire prefix.
Thus, \SLOT{} reuses the teacher signal required by OPD while enabling termination before uninformative suffix tokens are generated.

\begin{algorithm}[t]
\caption{\SLOT{}-OPD}
\label{alg:slot}
\begin{algorithmic}[1]
\Require warmup $K$, window $W$, trigger length $T$, buffer size $B$, percentile $\eta$, exemption $L_0$, max length $L_{\max}$.
\State Initialize FIFO buffer $\mathcal{B} \gets \emptyset$.
\For{optimizer step $k = 1, \dots, N$}
    \If{$k > K$ and $|\mathcal{B}|$ sufficient}
        \State $s_k \gets Q_{1-\eta}(\mathcal{B})$
    \EndIf
    \For{prompt $x$ in batch}
        \State $h_1 \gets x$;\quad $c \gets 0$;\quad $m \gets +\infty$
        \For{$t = 1, \dots, L_{\max}$}
            \State $\hat y_t \sim \student(\cdot \given h_t)$
            \State Compute (or estimate) $d_t = \KL(p_t \| q_t)$
            \If{$t \ge L_0 + W$}
                \State $z_t \gets \frac{1}{W} \sum_{i=t-W+1}^{t} d_i$
                \State $m \gets \min(m, z_t)$
                \If{$k > K$ and $z_t < s_k$}
                    \State $c \gets c + 1$
                \Else
                    \State $c \gets 0$
                \EndIf
                \If{$c \ge T$}
                    \State $\tau \gets t$;\quad \textbf{break}
                \EndIf
            \EndIf
        \EndFor
        \If{$m < +\infty$}
            \State push $m$ into $\mathcal{B}$
        \EndIf
    \EndFor
    \State Compute standard OPD update on $\{(t, \hat y_t) : t \le \tau\}$.
\EndFor
\end{algorithmic}
\end{algorithm}

\section{Implementation Details}
\label{sec:impl}

\paragraph{Models and data.}
All experiments use models from the Qwen3 family. The teacher is Qwen3-8B,\footnote{\url{https://huggingface.co/Qwen/Qwen3-8B}} and we consider two students of different scales: Qwen3-1.7B-Base\footnote{\url{https://huggingface.co/Qwen/Qwen3-1.7B-Base}} and Qwen3-4B-Base.\footnote{\url{https://huggingface.co/Qwen/Qwen3-4B-Base}} Training prompts are drawn from the DAPO-Math-17K dataset.\footnote{\url{https://huggingface.co/datasets/BytedTsinghua-SIA/DAPO-Math-17k}} For evaluation, we measure mathematical reasoning on four standard benchmarks: AMC,\footnote{\url{https://huggingface.co/datasets/AI-MO/aimo-validation-amc}} MATH500,\footnote{\url{https://huggingface.co/datasets/HuggingFaceH4/MATH-500}} MinervaMath,\footnote{\url{https://huggingface.co/datasets/math-ai/minervamath}} and AIME24.\footnote{\url{https://huggingface.co/datasets/Maxwell-Jia/AIME_2024}} All model checkpoints and datasets are obtained from the Hugging Face Hub.

\paragraph{Distillation objective.}
We run pure on-policy distillation ($\lambda\!=\!1.0$) with reverse KL and softmax temperature $\tau\!=\!1.0$. To control the cost of the divergence over a $\sim\!150\text{K}$-token vocabulary, both distributions are restricted to a truncated top-$K$ support ($K\!=\!64$) and re-normalized, with an explicit tail bucket carrying the residual probability mass; we find this approximation to match the full-vocabulary loss to within $\approx\!0.5\%$ relative error while reducing the loss-side FLOPs by more than two orders of magnitude.

\paragraph{Student rollouts.}
During training, the student generates completions via a colocated vLLM engine ($\text{TP}\!=\!1$, \texttt{gpu\_memory\_utilization} $=0.3$ for the 1.7B student and $0.2$ for the 4B student) under nucleus sampling with $T\!=\!1.0$, top-$p\!=\!0.95$, top-$k$ disabled, and one completion per prompt. Validation rollouts use an independent budget of $4096$ completion tokens. Student weights are synchronized into the vLLM engine after every optimizer step.

\paragraph{KL-based rollout truncation.}
Our proposed truncation operates on a sliding window of $L\!=\!64$ response-token reverse KL values, opened after the first $t\!=\!16$ tokens ($t\!=\!32$ for the 4B student). A FIFO buffer of $M\!=\!4096$ per-rollout minimum-window means is maintained on each worker; truncation is activated only after the buffer holds at least $512$ samples and after a warm-up of $K\!=\!50$ optimizer steps. The truncation threshold is set to the $(1-\eta)$-quantile of the buffer with $\eta\!=\!0.3$, and a rollout is cut once $N\!=\!4$ consecutive windows fall below this threshold. As an ablation we also evaluate a per-rollout uniformly-random truncation baseline with a minimum keep length of $1$ token.

\paragraph{Optimization.}
All experiments use AdamW with $\beta_1\!=\!0.9$, $\beta_2\!=\!0.999$, weight decay $0$, gradient clipping at $1.0$, and a cosine learning-rate schedule with $3\%$ linear warm-up. The peak learning rate is $1\!\times\!10^{-6}$ for on-policy distillation. We train the 1.7B student for $3$ epochs and the 4B student for $2$ epochs. Per-device train batch size is $1$ with $64$ gradient-accumulation steps, yielding an effective batch of $128$ sequences with $2$ GPUs (1.7B) and $256$ sequences with $4$ GPUs (4B). Dropout is disabled, and all reported numbers are averaged over three runs with different random seeds for reproducibility.

\paragraph{Systems and hardware.}
All experiments run on a single node with $4\!\times\!$NVIDIA A800-SXM4 $80$\,GB GPUs (NVLink). Training is launched through HuggingFace \texttt{Accelerate} in BF16 mixed precision with FlashAttention-2. The 1.7B student uses plain DDP across $2$ GPUs; the 4B student uses DeepSpeed ZeRO-2 across $4$ GPUs (no parameter or optimizer offload). Gradient checkpointing is enabled on the student; the teacher is kept frozen and runs a single forward pass per micro-batch.

\end{document}